# Neural Abstractive Text Summarizer for Telugu Language

Mohan Bharath B[1], Aravindh Gowtham B[2], Akhil M[3]

[1] M. Tech, IIIT Allahabad, Allahabad, U.P, India
[2] B. Tech, Shiv Nadar University, Greater Noida, U.P, India
[3] M. Tech, University of Hyderabad, Hyderabad, Telangana, India

[1] bharathhbm@gmail.com, [2] bg739@snu.edu.in, [3] akhil9351@gmail.com

**Abstract.** Abstractive Text Summarization is the process of constructing semantically relevant shorter sentences which captures the essence of the overall meaning of the source text. It is actually difficult and very time consuming for humans to summarize manually large documents of text. Much of work in abstractive text summarization is being done in English and almost no significant work has been reported in Telugu abstractive text summarization. So, we would like to propose an abstractive text summarization approach for Telugu language using Deep learning. In this paper we are proposing an abstractive text summarization Deep learning model for Telugu language. The proposed architecture is based on encoder-decoder sequential models with attention mechanism. We have applied this model on manually created dataset to generate a one sentence summary of the source text and have got good results measured qualitatively.

## 1 Introduction

Textual data is ever increasing in the current Internet age. We need some process to condense the text and simultaneously preserving the meaning of the source text. Text summarization is creating a short, accurate and semantically relevant summary of a given text. It would help in easy and fast retrieval of information. Text summarization can be classified into two categories.

- **Extractive Text Summarization** methods form summaries by copying from the parts of the source text by taking some measure of importance on the words of the source text and then joining those sentences together to form a summary of the source text.

- **Abstractive text summarization** methods create new semantically relevant phrases, it can also form summaries by rephrasing or by using the words that were not in the source text. Abstractive methods are actually harder . For an accurate and semantically relevant summaries, the model is expected to comprehend the meaning of the text and then try to express that understanding using the relevant words and phrases. So, abstractive models can have capabilities like generalization, paraphrasing.

Significant work is being focused on extractive text summarization methods and especially with English as the source language. There is no reported work for Telugu abstractive Text summarization using Deep learning models and also there are no available datasets for Telugu text summarization. Our goal is to build a model such that when given the telugu news article it should output semantically relevant sentence as the summary/title sentence for the corresponding telugu article. We have proposed a Deep learning model using encoder-decoder architecture and we have achieved good results measured qualitatively.

We have manually created the Dataset because of the fact there are no available datasets. Training Dataset has been created from the Telugu News websites by taking the headline as the summary and the main content as the source text and we have created a dataset with 2000 telugu news articles with their corresponding summaries which are taken as the headline of the respective article. We have created the dataset in such a way that the articles belonging to the different domains i.e, Politics, Entertainment, Sports, Business, National are more or less equally distributed to maintain a balance to the dataset.

To create word embeddings for the telugu words, we have made use of word-embeddings by FastText, which has created word embeddings for nearly 157 languages with each word-embedding of 300 dimensions.[1]

## 2 Related Work

As our Work is based on Abstractive Text Summarization using Deep learning models on Telugu Language on which there is no reported work. Our Deep Learning Models are mainly inspired by these three papers, Rush et al.[2]  used  an Encoder-Decoder Neural Attention Model to perform Abstractive Text Summarization on English text-data and found that it performed very well and beat the previous non Deep Learning-based approaches, Konstantin Lopyrev proposed a Encoder-Decoder recurrent neural network with attention mechanism to Generate Headlines for English News[3],and attention mechanism itself is inspired from Bahdanau seminal paper[4].

# 3 Approach

In this section we provide a brief overview of the Model/Architecture used and its individual components.

## 3.1 Recurrent Neural Network Encoder-Decoder

### 3.1.1 RNNs

Text is a sequential type of data. Recurrent Neural Networks are a type of Neural Networks used for handling sequential data. RNNs can take a variable length input sequence $X = (x_1, x_2, x_3, \ldots x_t)$ and can output sequence $Y = (y_1, y_2, y_3, \ldots, y_t)$. It uses an internal hidden state(h) to capture both the current input and the previous hidden state. A simple mathematical representation of recurrent network can be seen below.

$h_t = \sigma_h (W_h x_t + U_h h_{t-1} + b_h)$ (hidden state at timestep t)

$y_t = \sigma_y (W_y h_t + b_y)$       (output at time-step t)

$W_h$ is the weight matrix connecting input layer to hidden states and $U_h$ is the weight matrix connecting the hidden states of current time step to the hidden states of previous timestep. $b_h$ and $b_y$ are biases of hidden states and of the output layer. W, U and b are parameters and they are the same at each time step. $\sigma_h$ and $\sigma_y$ are activation functions. RNNs can also be multi layered they are called Deep RNNs, these are layered RNNs, where each layer extracts information from the previous layer.

### 3.1.2 LSTMs

Long short-term memory (LSTM) is a type of RNN architecture with complicated hidden unit computations. So, by introducing gates which are input, forget, output and memory cells, it allows memorizing and forgetting over a long distance of training and the model can effectively handle the vanishing gradient problem. LSTM is a basic unit of our Encoder-Decoder model to perform summarization.

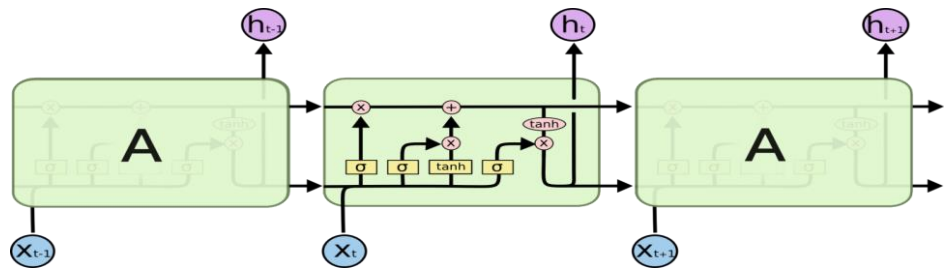

**Fig. 1.** LSTM UNIT

The motivation behind using an LSTM is that it captures long-term dependency pretty well and the information in the starting of the sequence is able to traverse down the line. This is done by being selective and restricting the information flow in the LSTM unit. There are three gates in an LSTM.

**Forget Gate Layer:**

- $f_t = \sigma(W_f [h_{t-1}; x_t] + b_f)$
- $C_t = C_t \odot f_t$

$C_t$ : Cell State at time step-t          $W_f$: weight matrix

$h_{t-1}$ : hidden state of previous timestep.          $X_t$ : input at timestep-t

$f_t$ : forget-gate          $\odot$ : Dot-product.

$\sigma$ : Sigmoid activation

**Input Gate Layer:**

- $I_t = \sigma(W_i[h_{t-1}; x_t] + b_i)$

- $C_t = \tanh(W_i[h_{t-1}; x_t] + b_c)$

- $C_t = C_t \odot i_t$

$i_t$ : input gate.
$C_t$ : Cell state at timestep t.
$W_i$ : weight matrix between previous state and current input.

**Output Gate Layer:**

- $O_t = \sigma(W_o[h_{t-1}; x_t] + b_o)$
- $h_t = O_t * \tanh(C_t)$

**3.1.3 Encoder-Decoder Model:**

Encoder-Decoder model is based on neural networks which aims at handling the mapping between the highly structured input and output. In the vanilla encoder-decoder model, the encoder RNN first reads the source text word by word and then encodes the information in a hidden state and passes the information forward. Decoder starts from the final hidden state of the encoder and at every timestep it computes a probability distribution on the total words in the vocabulary by taking a SoftMax function which gives probability values to all the words in the vocabulary and the most probable word is selected for that timestep and this continues until the end of sentence token is selected by the decoder or until the no of timesteps reaches the threshold. All the words generated so far will form the summary sentence!

- $x = (x_1, x_2, \ldots\ldots\ldots, x_T)$ be a length T input sequence to the encoder network.

- $y = (y_1, y_2, \ldots\ldots\ldots, y_U)$ be a length U output sequence the decoder network generates.

- Each Encoded representation $h_T$ contains information about the input sequence with focus on the $T^{th}$ input of the sequence.

- $h = (h_1, h_2, \ldots\ldots\ldots, h_T)$ be the encoder network output of length T.

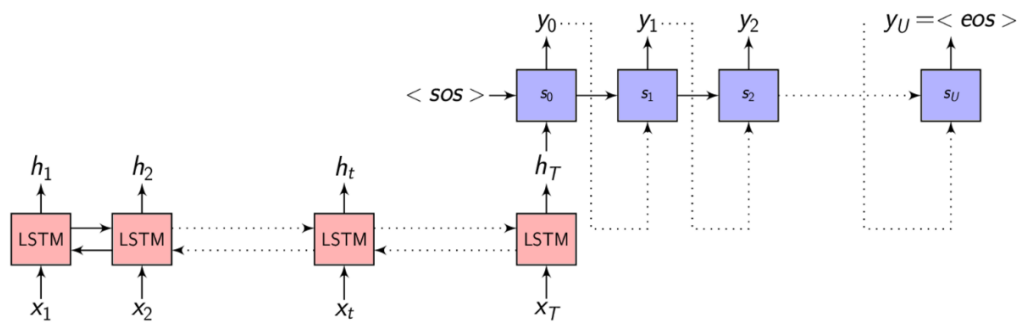

**Fig. 2.** Encoder-Decoder Architecture

In the Encoder-Decoder framework the encoder tries to summarize the entire input in a fixed dimension vector $h_t$. Decoder takes as input the output of the previous step and the hidden-state vector from the previous time step. So, at every timestep of the Decoder, the word selected in the previous time step and the hidden-state vector of the previous time step is given as input in the current timestep.

### 3.1.4 Attention Mechanism [4]

The basic Encoder-Decoder model performs well on very-short sentences but it fails to generalize well for longer sentences/paragraphs.

- The only input to the Decoder at the first timestep is a fixed size vector from the Encoder stage last timestep.
- This fixed size vector may not be able to capture all the relevant information of the longer source text.
- At each step of the decoder only certain parts of the input are relevant to generate an appropriate word.
- How do we make the model learn what to focus on at each step of the decoder? Attention!

Attention model calculates the importance of each input encoding for the current step of the Decoder by doing a kind of similarity check between decoder output at this timestep and all the input encodings. This similarity check is done by taking Dot product between the current hidden state of the decoder and all the input encodings. Doing this for all of the input encodings and normalizing, we get an importance Vector. We then convert it to probabilities by passing through SoftMax, which would give probability distribution. Then we form a context vector by multiplying with the encodings.

- importance$_{it}$ = V * tanh($e_i W_1 + h_t W_2 + b_{attn}$)
- Attention Distribution $a_t$ = SoftMax (importance$_{it}$)
- Context vector $h_t^*$ = $\sum e_i * a_i^t$

Context Vector is then fed into two layers to generate distribution over the vocabulary from which we sample.

- For the loss at time step t, loss$_t$ = -log P($w_t^*$), where $w_t^*$ is the target summary word. i.e. negative log probability of the target summary word.

- Loss at all the time steps is summed up.

- Now, we can do backpropagation and get all the required gradients and to minimize the loss we can apply gradient descent and learn the parameters of the network.

## 4. Training

Language: Python
Libraries: TensorFlow, Keras, etc.

- Dataset which is created manually, contains 2000 Telugu news articles and their corresponding summaries which are taken from the Headline of the respective articles.
- Dataset is divided into two parts: 1700 for Training and 300 for testing.
- Model has 128 hidden units.
- Loss function is cross-entropy.

- Training is done for 40 epochs.
- Training of Encoder-Decoder Architecture is done end to end i.e, loss is propagated to the encoder as well.

## 5. Evaluation

Evaluation of the results i.e, generated summaries of the source text have been done qualitatively. Here, a sample of the generated summaries has been given below.

**Table 1.** Results

| Source Text | Original Summary | Generated Summary |
|---|---|---|
| టిక్ టాక్ యాప్ ను భారతీయులు అధికంగా వినియోగిస్తున్నారు. రోజురోజుకు టిక్ టాక్ యాప్ ను ఉపయోగిస్తున్నవారి సంఖ్య పెరుగుతోంది. యూజర్లు తమ వీడియోల ద్వారా ప్రతిభను బయటపెడుతున్నారు. అదేవిధంగా చాలా మంది ఈ టిక్ టాక్ వీడియోల్లో డాన్స్ లు, పాటలు, చాలెంజ్ లు చేస్తూ.. సోషల్ మీడియాలో పాపులర్ అయిన విషయం తెలిసిందే. సాధారణంగా ఈ టిక్ టాక్ వీడియోలను చూస్తూ.. వైరల్ గా మారిన వీడియోలను షేర్ చేస్తూ యూజర్లు గంటల కొద్ది సమయాన్ని గడిపేస్తున్నారు. కొంతమందికి తమ ప్రతిభను బయట పెట్టడానికి.. మరికొంతమందికి కాలక్షేపం, వినోదానికి అనువుగా ఉండటంలో ఈ యాప్ పై యూజర్లు బోలెడంత సమయాన్ని కేటాయిస్తున్నారు. | భారతీయులు టిక్ టాక్ లో తెగ గడిపేశారు | భారతీయులు టిక్ టాక్ లో తెగ గడిపేశారు |
| సూర్యుడికి సంబంధించిన అత్యంత అరుదైన ఫొటోలను అమెరికా ఖగోళ శాస్త్రవేత్తలు విడుదల చేశారు. ప్రపంచంలోనే అత్యంత పెద్దదైన సోలార్ టెలిస్కోప్ గా ప్రసిద్ధి పొందిన డేనియల్ కే ఇనౌయే సోలార్ టెలిస్కోప్ (డీకేఐఎస్ టీ) | ముందెన్నడూ చూడని సూర్యుడి అద్భుత ఫొటోలు! | ముందెన్నడూ చూడని సూర్యుడి అద్భుత ఫొటోలు! సూర్యుడి అద్భుత |

అద్భుత ఆవిష్కారానికి కారణమైంది. దీని ద్వారా సూర్యుడి ఉపరితలానికి సంబంధించిన అరుదైన ఫొటోలను చూసే అవకాశం మానవాళికి దక్కింది. కాగా హవాయి ద్వీపంలో ఏర్పాటు చేసిన ఈ భారీ టెలిస్కోపు ద్వారా సూర్యుడిని అత్యంత సమీపంగా చూస్తూ.. అంతర్ధత శక్తిని అంచనా వేసే అవకాశం ఉంటుందని ఆస్ట్రోనాట్లు పేర్కొంటున్నారు. ప్రస్తుతం ఇది విడుదల చేసిన ఫొటోల ఆధారంగా.. సూర్యుడి ఉపరితలం మీద కణాల వంటి ఆకారాలను జూమ్ చేయగా.. ఒక్కోటి అమెరికా రాష్ట్రం టెక్సాస్ పరిమాణంలో ఉందని తెలిపారు.

ఒలింపిక్ ఏడాది నెపథ్యంలో భారత బ్యాడ్మింటన్ సింగిల్స్ కోచ్ గా సింగిల్స్ కోచ్ గా ఇండోనేసియాకు చెందిన అగుస్ డ్వి సాంటొసొ సాంటోసొను ఎంపిక చేస్తూ కేంద్ర క్రీడా మంత్రిత్వ శాఖ నిర్ణయం తీసుకుంది. దక్షిణ కొరియాకు చెందిన కిమ్ జి హ్యూన్ వెళ్లడంతో ఏర్పడిన కోచ్ ఖాళిని భర్తీ చేయాలంటూ స్పోర్ట్స్ అథారిటీ ఆఫ్ ఇండియా (సాయ్ ) గతంలో కేంద్ర క్రీడా మంత్రిత్వ శాఖను కోరింది. దానిపై స్పందించిన మంత్రిత్వ శాఖ సాంటోస్ ను నియమిస్తూ నిర్ణయం తీసుకుంది. అతడు ఒలింపిక్స్ ముగిసే వరకు కోచ్ గా సేవలు అందించనున్నాడు. సాంటోస్ మార్చి రెండో వారంలో భారత బ్యాడ్మింటన్ జట్టుతో కలుస్తాడు. అతడి పర్యవేక్షణలో ప్రపంచ ఛాంపియన్ పీవీ సింధుతో పాటు ఇతర సింగిల్స్ పట్టర్లు కూడా టోక్యో కోసం సిద్ధమవుతారు. సాంటోస్ ఇకఛనల్ సంతృప్తి చెందితే అతడిని 2024 వరకు కూడా కొనసాగిస్తామని భారత బ్యాడ్మింటన్ సంఘం (బాయ్ ) కార్యదర్శి అజయ్ సింహానియ తెలిపారు. ఒలింపిక్స్ వరకు సాంటోసకు నెలకు 8 వేల డాలర్లు (సుమారు రూ.5.8 లక్షల ) చెల్లించనున్నారు.

అనుమానాస్పద కార్యకలాపాలను కొనసాగించే మొబైల్ అప్లికేషన్లను గుర్తించి వాటికి చెక్ పెట్టే అదునాతన ఫీచర్ ను పేటీఎం పేమెంట్స్ బ్యాంక్ అందుబాటులోకి తీసుకొచ్చింది. 'రోగ్ ' పేరిట ఈ ఫీచర్ ను అందిస్తోంది. మోసపూరిత లావాదేవీలను పసిగట్టి.. ఏ యాప్ ద్వారా సమాచారం చోరింఛో తెలుసుకుని, అటువంటి యాప్ లను గుర్తించి వాటిని అన్ ఇన్ స్టాల్ చేయమని వినియోగదారులకు సూచిస్తుంది.

మోసపూరిత యాప్ లకు పేటీఎం చెక్

దేశంలో తొలి ఎలక్ట్రిక్ ట్రాక్టర్ ను హైదరాబాద్ కు చెందిన స్టార్టప్ సెలెస్టియల్ ఈ–మొబిలిటీ రూపొందించింది. వినియోగానికి వీలున్న నమూనాను బుధవారమిక్కడ ఆవిష్కరించింది. ఉద్యానవనాలు, విమానాశ్రయాలు, ఫ్యాక్టరీలు, గిడ్డంగుల్లో సరుకు రవాణాకు వీలుగా 6 హెచ్ పీ సామర్థ్యంతో తయారు చేశారు. 21 హెచ్ పీ డిజెల్ ట్రాక్టర్రుకు సమానంగా ఇది పనిచేస్తుందని కంపెనీ సహ వ్యవస్థాపకుడు సిద్ధార్థ దురైరాజన్ మీడియాకు తెలిపారు. 'ధర రూ.5 లక్షల లోపు ఉంటుంది. ప్రభుత్వం నుంచి సబ్సిడీ కూడా అందుకోవచ్చు. డిజెల్ ట్రాక్టరుతో గంటకు రూ.150 ఖర్చు వస్తే, దీనికి రూ.20–35 మధ్య ఉంటుంది. ఒకసారి చార్జింగ్ చేస్తే 75 కిలోమీటర్లు ప్రయాణిస్తుంది. వేగం గంటకు 20 కిలోమీటర్లు. 5–8 ఏళ్లు బ్యాటరీ మన్నికగా ఉంటుంది. నెలకు 100 ట్రాక్టర్ల తయారీ సామర్థ్యంతో బాలానగర్ లో ఫ్యాక్టరీ ఉంది. రూ.60 కోట్ల దాకా నిధులు సమకరించనున్నాం' అని వివరించారు.

బారత్ లో తొలి ఎలక్ట్రిక్ ట్రాక్టర్

ఈ ఏడాది దేశీయ రియల్టీ రంగం ఆశించనంత వృద్ధిని సాధించలేదు. వినియోగ వ్యయం తగ్గడం, పెట్టుబడులు క్షీణించ డం, ప్రపంచవ్యాప్తంగా ఆర్థిక మందగమనం వంటి రకరకాల కారణాలతో దేశీయ రియల్టీ రంగం లో వృద్ధి అవకాశాలను నిరుగార్చాయని అనరాక్ ప్రాపర్టీ కన్సల్టెంట్స్ తెలిపింది. దేశంలోని ఏడు ప్రధాన నగరాల్లో గతేడాది నాలుగు త్రైమాసికాలు కలిపి 2,48,300 గృహాలు అమ్ముడుపోగా.. 2019లో కేవలం 4 శాతం

రియల్టీ రంగానికి 2019లో నిరాశ

వృద్ధితో 2,58,410 యూనిట్లకు చేరాయి. ఇందులోనూ అందుబాటు గృహాల విక్రయాలే ఎక్కువగా ఉన్నాయి. అపడంబు ల్ హౌసింగ్ లకు పలు పన్ను రాయితీలను కల్పించ డమే ఇందుకు కారణం. తొలిసారి గృహ కొనుగోలుదారులకు రూ.3.5 లక్షల పన్ను రాయితీని అందిస్తుంది. ఇది 2020 ఆర్థిక సంవత్సరం ముగిసె వరకూ అందుబాటులో ఉంటుంది.

వారానికి మూడు సార్లు చేపను ఆహారంగా తీసుకుంటే క్యాన్సర్ ముప్పు గణనీయంగా తగ్గుతుందని తాజా అధ్యయనం వెల్లడించింది. వారానికి ఒకసారి చేపను తినేవారితో పోల్చితే మూడు సార్లు తీసుకునేవారిలో పేగు క్యాన్సర్ ముప్పు 12 శాతం తక్కువగా ఉందని ఈ పరిశోధన వెల్లడించింది. అన్ని రకాల చేపలను తీసుకోవడం మంచిదే అయినా నూనె అధికంగా ఉండే సాల్మన్ , మాకరెల్ చేపల కంటె ఇతర చేపలు మరింతగా ఆరోగ్య ప్రయోజనాలు అందిస్తాయని తెలింది.

చేపలతో క్యాన్సర్ కు చేపలతో కేక్ తినడం వల్లె చెక్ తండ్రికొడుకులు చెక్

---

We have observed that the generated summaries are more or less semantically relevant to the source text in many cases in the test data, despite the limited training data our model did a good job.

## 6 Conclusions

We have implemented abstractive text summarization for Telugu language using encoder-decoder architecture with attention mechanism. Specifically, there are no datasets for Telugu language that have paired human-generated summaries. We have got semantically relevant good results on most of the test data measured qualitatively, although there are very few generated summaries which were not at all relevant. Given the fact the Dataset has been created from the scratch and the limitation of the dataset size, we have got good results. For text summarization in Telugu language, one of the difficulties is the lack of quality summaries of large given dataset, in future we would like to create a very large dataset corpus for this purpose comparable to the standard English text summarization datasets available, we would also work towards a numeric metric specifically to telugu text summarization tasks which would also capture the semantic relevance of the summary to the given source text .In future we shall explore the possibility of using the transformer architecture for telugu text summarization task.

We have taken permission from competent authorities to use the images/data as given in the paper. In case of any dispute in the future, we shall be wholly responsible.

# References

1. Edouard Grave, Piotr Bojanowski, Prakhar Gupta, Armand Joulin, Tomas Mikolov. "Learning Word Vectors for 157 Languages."arXiv:1802.06893v2 [cs.CL]
2. Alexander M. Rush, Sumit Chopra, Jason Weston, "A Neural Attention Model for Abstractive Sentence Summarization." arXiv:1509.00685v2 [cs.CL]
3. Konstantin Lopyrev, "Generating News Headlines with Recurrent Neural Networks" arXiv:1512.01712v1[cs.CL]
4. Dzmitry Bahdanau, Kyughyun Cho, Yoshua Bengio Neural Machine Translation by Jointly Learning to Align and Translate. arXiv:1409.0473v7 [cs.CL]